\documentclass[10pt,twocolumn,letterpaper]{article}
\pdfoutput=1
 
\usepackage{cvpr}
\usepackage{times}
\usepackage{epsfig}
\usepackage{graphicx}
\usepackage{amsmath}
\usepackage{amssymb}
\usepackage{enumitem}

\cvprfinalcopy 

\def\cvprPaperID{1289} 

\ifcvprfinal\fi
\newcommand{\bfx}{{\bf x}}

\newcommand{\bfy}{{\bf y}}

\newcommand{\bfe}{{\bf e}}
\newcommand{\bfA}{{\bf A}}

\newcommand{\reals}{{\mathbb R}}

\newcommand{\note}[1]{{\bf [{\em Note:} #1]}}

\newcommand{\Section}[1]{\vspace{-4pt}\section{#1}\vspace{-4pt}}

\newenvironment{tight_itemize}{\begin{itemize}[leftmargin=*] \itemsep
-3pt}{\end{itemize}}

\newcommand{\hgcnote}[1]{}
\newcommand{\Salnote}[1]{}

\graphicspath{{Figures/}}

\begin{document}

\title{FPA-CS: Focal Plane Array-based Compressive Imaging in  Short-wave Infrared}

\author{Huaijin Chen$^\dagger$, M. Salman Asif$^\dagger$, Aswin C.\ Sankaranarayanan$^\ddag$, Ashok Veeraraghavan$^\dagger$\\
$^\dagger$ECE Department, Rice University, Houston, TX\\
$^\ddag$ECE Department, Carnegie Mellon University, Pittsburgh, PA%
}

\maketitle

\begin{abstract}
Cameras for imaging in short and mid-wave infrared spectra are significantly more expensive than their counterparts in visible imaging.
As a result, high-resolution imaging in those spectrum remains beyond the reach of most consumers.
%
Over the last decade, compressive sensing (CS) has emerged as a potential means to realize inexpensive short-wave infrared cameras.
One approach for doing this is the single-pixel camera (SPC) where a single detector acquires coded measurements of a high-resolution image.
A computational reconstruction algorithm is then used to recover the image from these coded measurements.
Unfortunately, the measurement rate of a SPC is insufficient to enable imaging at high spatial and temporal resolutions.

We present a focal plane array-based compressive sensing (FPA-CS) architecture that achieves high spatial and temporal resolutions.
The idea is to use an array of SPCs that sense in parallel to increase the measurement rate, and consequently, the achievable spatio-temporal resolution of the camera.
%
%
%
We develop a proof-of-concept prototype in the short-wave infrared using a sensor with 64$\times$ 64 pixels; the prototype provides a 4096$\times$ increase in the measurement rate compared to the SPC and achieves a megapixel resolution at video rate using CS techniques.
\end{abstract}

\section{Introduction}
The cost of a high-resolution sensors in the visible spectrum has fallen dramatically over the last decade. 
For example, a  cellphone camera module boasting a sensor with several megapixels costs little more than a few dollars.
This trend is fueled by the fact that silicon is sensitive to the visible region of the electromagnetic spectrum and hence, the scaling trends and advances made in silicon-based semiconductor fabrication  directly benefit visible imaging technologies.
Unfortunately, these scaling trends do not extend to imaging beyond the visible spectrum. 

\begin{figure}[t]
  \centering
  	\includegraphics[width = 0.45\textwidth]{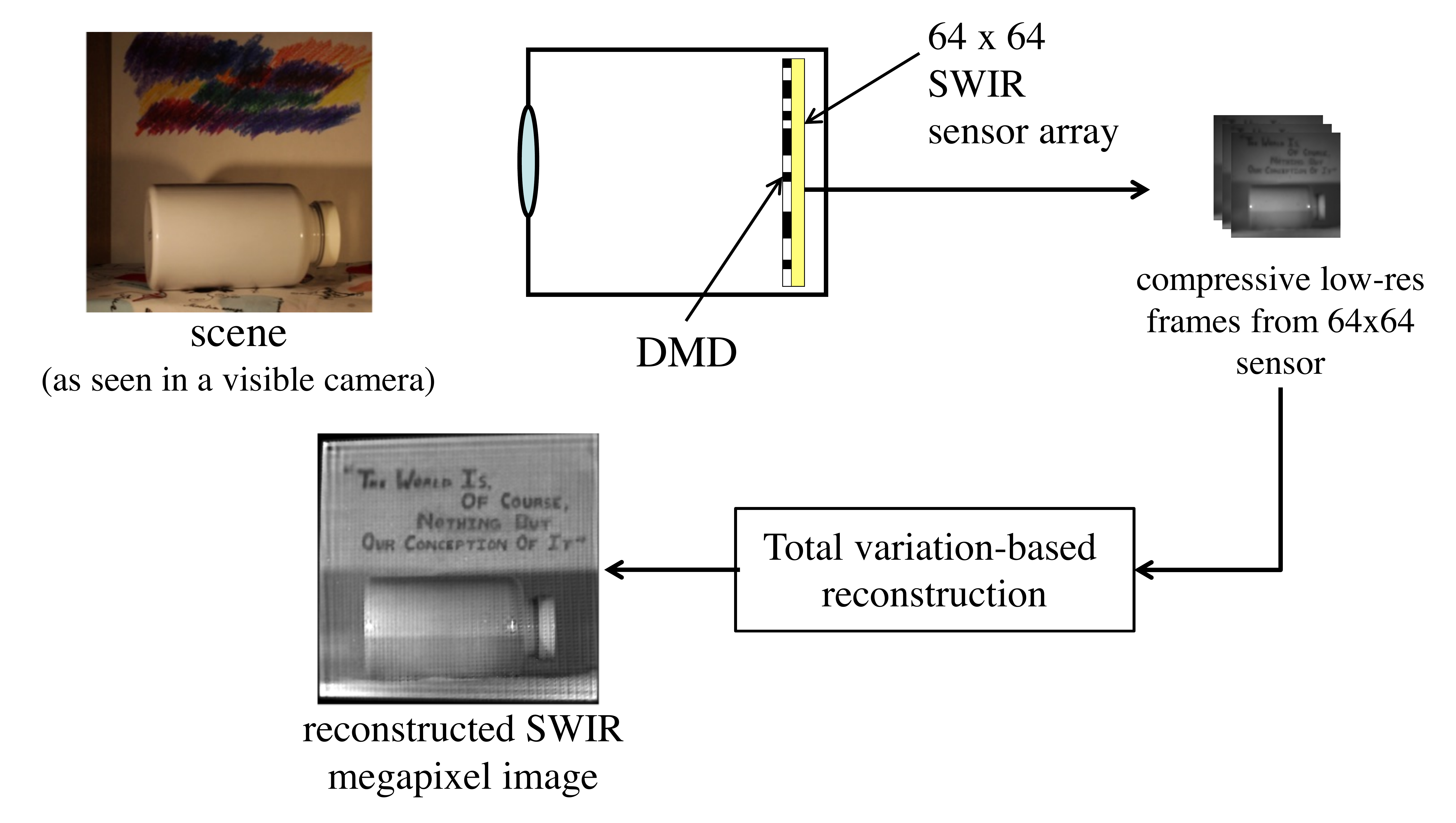}
    \caption{Focal plane array-based compressive sensing (FPA-CS) camera with a $64 \times 64$ SWIR sensor array is equivalent to $4096$ single pixel cameras (SPCs) operating in parallel. This results in vastly superior spatio-temporal resolutions against what is achievable using the SPC or a traditional camera.}
    \vspace{-4mm}
    \label{fig:overview}
\end{figure}

%

\vspace{-4mm}
\paragraph{Motivation.}
In many application domains, imaging beyond the visible spectrum provides significant benefits over traditional visible sensors \cite{hansen2008overview}.
For example, short-wave infrared (SWIR)  penetrates fog and smog; this enables imaging through  scattering media.
The night-glow of the sky naturally provides SWIR illumination  which enables SWIR sensors to passively image even in the dark.
%
%
SWIR imaging also enables a wide variety of biomedical applications \cite{schmitt1999optical}.
Yet,  SWIR imaging requires sensors made of exotic materials such as indium gallium arsenide (InGaAs), which are orders of magnitude more expensive than silicon.
As a consequence, the cost of a megapixel sensor in the SWIR regime is still greater than tens of thousands dollars. 
Hence, despite their immense potential, high-resolution SWIR cameras are beyond the reach of engineers and scientists in application domains that could most benefit from its use.

In this paper, we leverage the theory and practice of compressive sensing (CS) \cite{candes2006robust, donoho2006compressed}, to enable high-resolution SWIR imaging from low-resolution sensor arrays.
%
%
CS relies on the ability to obtain arbitrary linear measurements of the scene; this requires a fundamental redesign of the  architecture used to image the scene.
The single-pixel camera (SPC) is an example of such an architecture \cite{duarte2008single}. 
The SPC uses a digital micro-mirror device (DMD) as a spatial light modulator and acquires coded measurements of an image onto a single photo-detector. 
We can build an SWIR SPC by employing a photo-detector sensitive to SWIR along with the appropriate choice of optical accessories.
The programmable nature of the DMD enables the photo-detector to obtain the sum of any arbitrary subset of pixels.
%
The high-resolution image can then be computationally reconstructed from a small number of such measurements.

A SPC is incapable of producing high-resolution images at video rate.
To understand why, we first observe that the measurement rate of an SPC is determined by the operating speed of its DMD which seldom goes beyond 20 kHz.
At this measurement rate, conventional Nyquist-based sampling with a single-pixel can barely support  a 20 fps video at a spatial resolution of $32\times 32$ pixels.
%
%
To sense at a mega-pixel resolution and video rate using a CS-based SPC, we would need a measurement rate of tens of millions of measurements per second --- a gap of 1000$\times$ that is beyond the capabilities of current CS techniques.
%

\vspace{-4mm}
\paragraph{Applications of SWIR Imaging.} A large number of applications that are difficult or impossible to perform using the visible spectrum become much simpler due to the characteristics of the SWIR spectrum.
SWIR imaging is currently used in a host of applications including automotive, electronic board inspection, solar cell inspection, produce inspection, identification and sorting, surveillance, anti-counterfeiting, process quality control, and much more \cite{hansen2008overview}. 
Some unique properties of SWIR that enable many of these applications include (a) improved penetration through scattering media including tissue, fog, and haze, (b) seeing through many commonly-used  packaging materials which are transparent in SWIR while opaque in visible spectrum, and (c) observing defects and gauging quality of fruits and vegetables. Finally, for night-time surveillance application, the night-glow of the sky provides sufficient SWIR illumination even on a moon-less night; this enables long-distance SWIR imaging without the need for extra illumination sources that could compromise reconnaissance.

\vspace{-4mm}
\paragraph{Contributions.}
This paper enables a novel class of CS architectures that achieve high spatial and temporal resolutions  using inexpensive low-resolution sensors.
The main technical contributions in this paper are:
\begin{tight_itemize}
\item{We  characterize the spatio-temporal resolution limits of CS architectures. A key finding is that  a space-bandwidth product mismatch between the DMD and the photo-detector results in sub-optimal performance.}
\item{We propose the focal plane array-based compressive sensing (FPA-CS) camera---an imaging architecture that is optically identical to thousands of SPCs acquiring compressive measurements in parallel (see Figure~\ref{fig:overview}). FPA-CS balances space-bandwidth product constraints, thereby enabling CS-based imaging architectures with higher spatial and time resolutions.}
\item{We develop a prototype FPA-CS camera in SWIR and demonstrate capturing 1 megapixel images at video rate, far exceeding the capabilities of current methods.}
\end{tight_itemize}

%

\Section{Related work}

\paragraph{Compressive sensing (CS).} Compressive sensing \cite{candes2006robust, donoho2006compressed} deals with the estimation of a  signal {$\bf x \in \reals^N$} from $M < N$ linear measurements $\bfy \in \reals^M$ of the form
\begin{equation} \bfy = A \bfx + \bfe, \label{eq:csmodel} \end{equation}
where $\bfe$ is the measurement noise and $A$ is the measurement matrix.
Estimating the signal $\bf x$ from the compressive measurements $\bf y$ is an ill-posed problem since the system of equations is under-determined.
Nevertheless, a fundamental result from CS theory states that a robust estimate of the vector $\bf x$ can be obtained from  $M \sim K\log(N/K)$
measurements if
the signal $\bf x$ admits a $K$-sparse representation and 
the sensing matrix $A$ satisfies the so-called restricted isometry property \cite{candes2008restricted}.
Furthermore, signals with sparse transform-domain coefficients or sparse gradients  can be estimated stably from the noisy measurement $\bfy$ by solving a convex problem \cite{candes2006robust,osher2005iterative}.
%

\begin{figure}[t]
  \centering
  	\includegraphics[width = 0.475\textwidth]{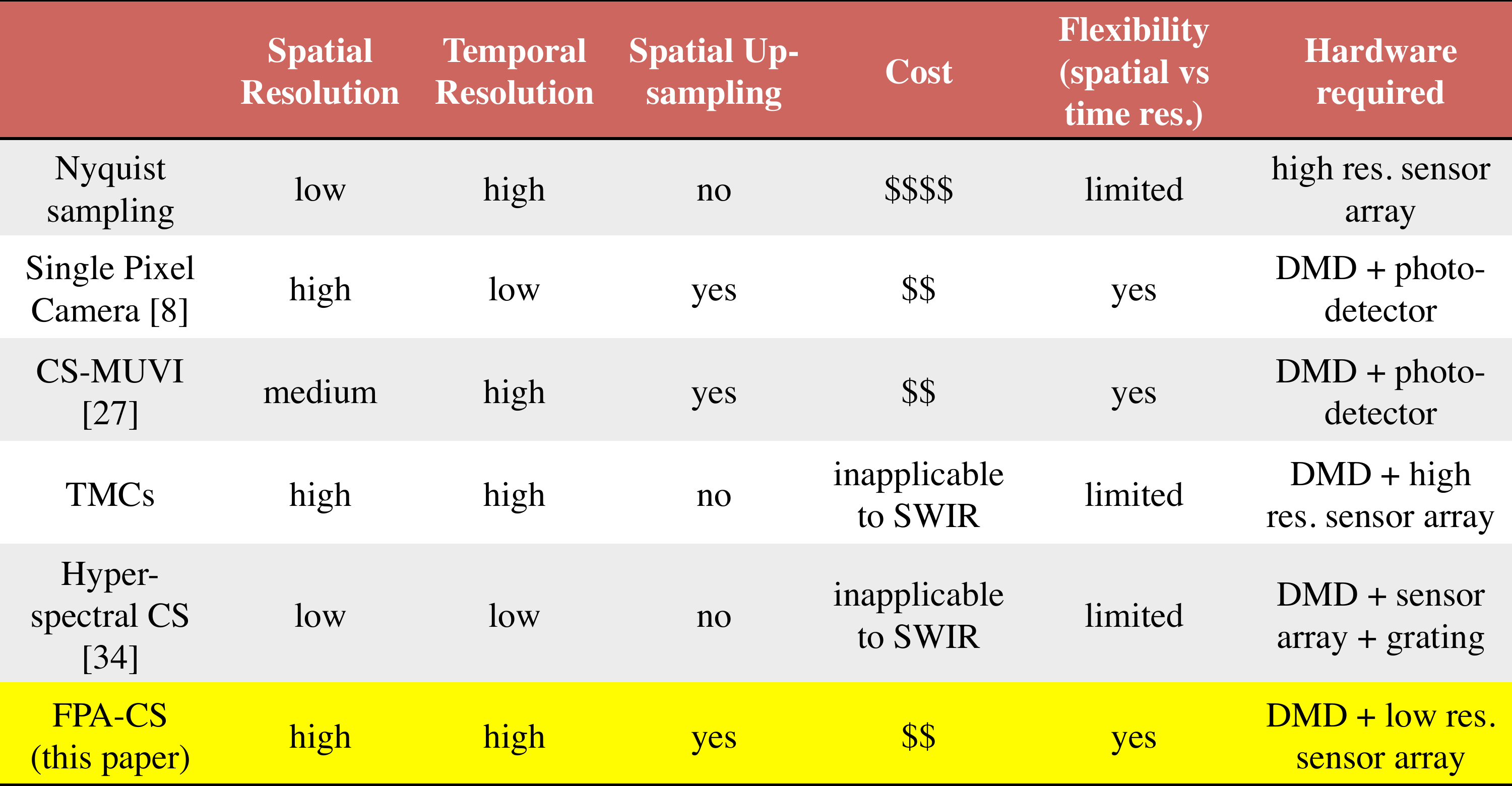}
    \caption{Comparisons of some existing CS-based cameras.}
    \label{fig:Related}
\end{figure}

\vspace{-4mm}
\paragraph{Compressive imaging architectures.}
In the context of video CS, there are two broad classes of  architectures: spatial multiplexing and temporal multiplexing cameras. Figure \ref{fig:Related} provides a comparison of various CS architectures 

\textit{Spatial multiplexing cameras (SMCs)}  acquire coded, low-resolution images and super-resolve them to obtain high-resolution images.
In particular, they employ a spatial light modulator (SLM), e.g., a digital micro-mirror device (DMD) or liquid crystal on silicon (LCoS), to optically compute linear projections of the scene~$\bf x$; these linear projections determine the rows of the sensing matrix $A$ in \eqref{eq:csmodel}. 
%
%
Since SMCs are usually built with only low-resolution sensors, they can operate at wavelengths where  full-frame sensors are too expensive.
%

A prominent example of SMC is the single pixel camera (SPC)~\cite{duarte2008single};  
its main feature is to sense using only a \emph{single} sensor element (i.e., a single pixel) and that the number of multiplexed measurements required for image reconstruction is significantly smaller than the number of pixels in the scene.
In the SPC, light from the scene is focused onto a programmable DMD, which directs  light from only a subset of activated micro-mirrors  onto the photodetector. 
%
By changing the micro-mirror configurations, we can obtain  linear measurements corresponding to the sensing model in (\ref{eq:csmodel}).
Several multi-pixel extensions to the SPC have been proposed recently, with the goal of increasing the measurement rate  \cite{ke2012object,kerviche2014information,mahalanobis2014recent,wang2015lisens}.   
To our knowledge, ours is the only design that focuses on sensing in SWIR wavebands.

A commercial version of the SPC for sensing in SWIR has been produced by InView Corporation. A key difference between our proposed architecture and the InView camera is the number of sensing elements. To our knowledge, the InView camera, much like the SPC, uses a single photo diode, whereas we use a sensor with $64 \times 64$ pixels. 

%
%


SMCs  for video CS also make use of a diverse set of signal models and constraints including  3D wavelets \cite{wakin2006compressive}, multi-scale wavelet lifting \cite{park2013multiscale}, optical flow-based reconstructions \cite{sankaranarayanan2012cs,AFR_videoCS_Asilomar}, block-based models \cite{fowler2010block}, sparse frame-to-frame residuals \cite{vaswani2010modified, cevher2008compressive}, linear dynamical systems \cite{vaswani2008kalman, sankaranarayanan2013compressive}, and combinations of low-rank and sparse matrices \cite{waters2011sparcs}.
One characteristic of all these algorithms is that reconstruction performance improves with increasing number of measurements. However, the measurement rate in traditional SPC architectures is too low to support high resolution, high frame rate reconstructions.
%

In sharp contrast to SMCs, \textit{temporal multiplexing cameras} (TMCs) use full-frame sensors with low frame rates and aim to super-resolve videos temporally, i.e., produce high frame rate videos from low frame rate sensors.
Veeraraghavan et al.~\cite{veeraraghavan2011coded} showed that periodic scenes could be imaged at high temporal resolutions by using global temporal coding. 
This idea was extended to non-periodic scenes in \cite{holloway2012flutter} where a union-of-subspace models was used.
Per-pixel temporal modulation to recover higher frame-rates was demonstrated using prototypes that used LCOS for modulation \cite{gupta2010flexible,reddy2011p2c2,hitomi2011video}.
Llull et al.\ \cite{llull2013coded} propose a TMC that uses a translating mask in the sensor plane to achieve temporal multiplexing. 
%
%
%
%
Harmany et al.\ \cite{harmany2013compressive} extend coded aperture systems by incorporating a flutter shutter \cite{raskar2006coded}; the resuling TMC provides immense flexibility in the choice of measurement matrix.
A common feature for all TMCs is the use of a high resolution sensor; this makes them inapplicable for SWIR imaging where high resolution sensor arrays are prohibitively expensive.
%


\vspace{-4mm}
\paragraph{Super-resolution (SR).}
SR is a technique that is commonly used to enhance the resolution of a given image. Traditional SR works by utilizing image priors or acquiring multiple measurements of the same scene \cite{park2003super}. However, traditional SR cannot arbitrarily increase resolution, and even state-of-the-art SR algorithms are limited to 2--4$\times$ upsampling \cite{glasner2009super}.
In contrast, for a loss in temporal resolution, our proposed architecture can achieve the full spatial resolution of the DMD without compression (see Section \ref{sec:expts} and Figures \ref{fig:resolution_charts}(f), \ref{fig:Invisible}(b)) with the sensor pixel-wisely scanning of the DMD, which can be seen as implementing SR in the optical domain: at each sensor pixel, we take 256 non-overlapping sub-pixel measurements by turning on different DMD pixels that map to the given sensor pixel. If we were to procure a DMD with higher resolution, the camera system inherits that same resolution. 

\section{Spatio-temporal resolution (STR)}

\paragraph{Nyquist cameras STR.}
The STR of a camera is limited by the product of the number of pixels and the maximum frame rate.
For example, a 1 megapixel sensor operating at 30\ fps provides a measurement rate, which we denote as $M_r$, equals to $30\times10^6$ samples per second.
Traditional cameras rely on the principle of Nyquist sampling; thus, for such cameras $\textrm{STR} = M_r$.
Pixel count and frame rate of visible sensors have significantly improved, and it is now common to obtain sensors that can achieve megapixel resolution at $30$ fps.
Unfortunately sensors outside the visible spectrum either provide a much lower spatial resolution, or they are quite expensive.

\vspace{-4mm}
\paragraph{Compressive cameras STR.}
%
%
Let us consider a CS-based camera operating at measurement rate, $M_r$ samples per second, that can provide a high-resolution video with $\alpha M_r$ pixels per seconds. The effective STR can be written as $\textrm{STR} = \alpha M_r$, where $\alpha \ge 1$ represents the compression factor by which the sampling rate reduced thank to the compressive sensing framework. 
In a SPC, the measurement rate is typically limited by the maximum rate at which synchronization can be achieved between the DMD modulator and the sensor.
While a photo-detector can be operated at very high rates (even GHz), commercially-available DMD seldom operate faster than  10--20~kHz.
Hence, it is not possible to achieve synchronization between any of the current high resolution spatial light modulators and a photo-detector at greater than $f_{DMD} = 20$ kHz.
This directly imposes a limit on the STR of compressive cameras based on single pixel sensors, i.e., $\textrm{STR} \leq  \alpha M_r = \alpha  f_{DMD}$ samples per second.


\vspace{-4mm}
\paragraph{Increasing the measurement rate.}
From the previous discussion, it is clear that in order to increase the STR of CS-based imaging systems, one must increase the measurement rate.
Given that the operating speed of the DMD poses strict limits on the number of frames we can obtain in unit time, one approach is to increase the measurement rate by reading multiple measurements in parallel.
As an example,  a compressive imaging system, in which a $K \times K$ pixel image sensor array is used to acquire multiplexed measurements in synchronization with a DMD at an operational rate $f_{DMD}$ Hz, provides a measurement rate of $M_r = K^2f_{DMD}$ samples per second---a $K^2$ times improvement over the SPC.
This increased measurement rate enables the acquisition of videos at higher spatial and temporal resolution.
In the next section, we describe a SWIR prototype that uses a $64 \times 64$ focal plane array sensor along with a DMD operating at $f_{DMD}=480$\ Hz to achieve measurement rates in millions of samples per second.

\section{Specifics of the FPA-CS prototype}
\label{sec:arch}
\begin{figure}[t]
	\centering
	\includegraphics[width = 0.5\textwidth]{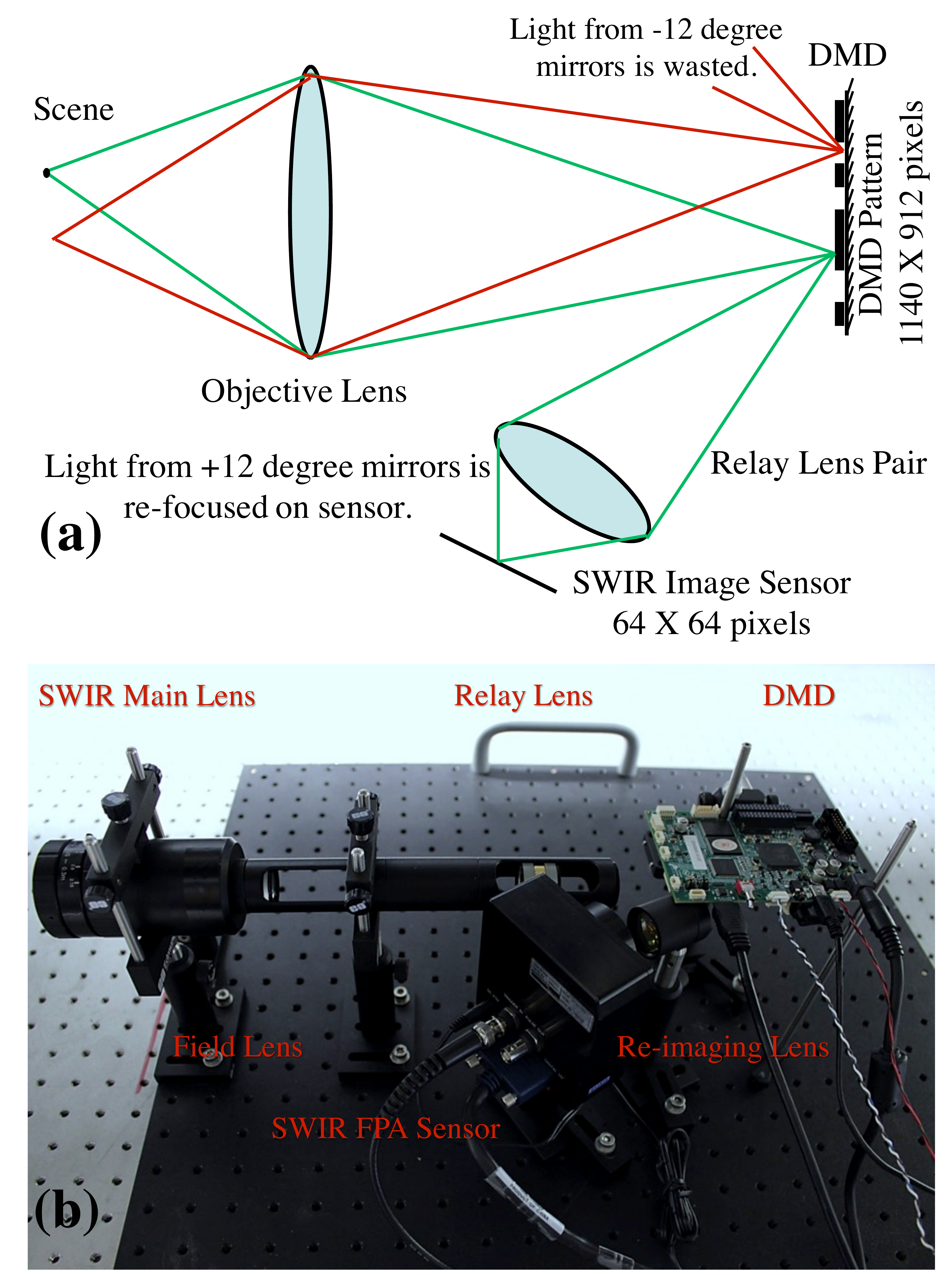}
	\caption{(a) Schematic illustration of the optical system. (b) Photograph of our prototype.}
	\vspace{-0.2in}
	\label{fig:prototype}
\end{figure}

\paragraph{System Architecture.} 
Figure \ref{fig:prototype} shows a schematic of our design and a photograph of our prototype.
We utilized a Texas Instruments Light Crafter DMD as the light modulator.
The DMD consists of $1140 \times 912$ micro-mirrors, each of size  $7.6$ micron. 
Each mirror can be independently rotated to either $+12 ^{\circ}$ or $-12 ^{\circ}$ around the optical axis at a rate of 2.88\ KHz.
We used a SWIR objective lens (Edmund Optics $\#83-165$) with a focal length of 50\ mm to focus the scene on to the DMD.
A 150\ mm-150\ mm relay lens pair (2x Thorlabs AC254-150C) was placed after the SWIR objective lens to extend the original flange distance, thereby providing ample space for the light bundle reflecting out of the DMD \Salnote{George: What is this? is this out of DMD or into the DMD}.
\hgcnote{out. how can I state it more clearly? I want to say that since the flange distance is longer, the out going ray bundle can be of a bigger angle, thus won't hit the end of objective lens.}  
We also used a 50\ mm field lens (Thorlabs LB1417) to reduce vignetting.
The light incident on the DMD corresponding to pixels that are oriented at $-12 ^{\circ}$ is discarded, while the light that is reflected from pixels that are oriented at $+12 ^{\circ}$ is focused on the SWIR sensor using a re-imaging lens.
We used a 100\ mm-45\ mm lens pair (Thorlabs AC-254-100C and AC-254-045C) as our re-imaging lens, which provides a 1:2.22 magnification as the physical sizes of the sensor and the DMD are different.
%

We used a $64 \times 64$ SWIR sensor (Hamamatsu G11097-0606S), with $50$ micron pixel size.
The relay lens is configured such that roughly $16 \times 16$ pixels on the DMD map to one sensor pixel.
Since the DMD has a rectangular shape and the sensor has a square one, some of the DMD pixels are not mapped on the sensor. In our prototype, the sensor measures a square region of the DMD with approximately 600,000 micromirrors, which we up-sample to a megapixel image. 
\Salnote{Let's talk about this 600K vs 1M resolution...}
Furthermore, since the DMD and the sensor planes are not parallel, we adjusted the relay lens and the sensor position to satisfy the Scheimpflug principle so that the entire scene plane remains in focus on the sensor~\cite{nayar2006programmable}.

\vspace{-4mm}
\paragraph{Achievable measurement rates.}
In our prototype, we operate the DMD at $f_{DMD} = 480$ Hz in synchronization with the readout timing of the $64\times 64$ sensor array. Therefore, we obtain the measurement rate, $M_r \approx 2 \times 10^6$ pixels/sec.
In our experimental results, we demonstrate high-fidelity video reconstruction at a compression rate of $\alpha = 16$. 
Thus, the effective  $\textrm{STR}$ of our system is limited by STR $ \le \alpha K^2 f_{DMD} = 32 \times 10^6$ pixels/second, which makes recovery of videos at spatial resolution of 1 megapixel at approximately 32 frames/sec possible. 
In comparison, a SPC with the DMD operating at 20,000 Hz and the compression factor $\alpha = 16$ would provide $1$ megapixel at $0.32 fps$. 
Clearly, the mismatch between the measurement rate of the sensor and the operating speed  of the DMD is the major bottleneck in SPC. 

It is worth noting here that an equivalent full-frame SWIR sensor that can operate at 1 megapixel spatial resolution over 30 fps would cost upwards of \$60,000. 
In contrast, our $64 \times 64$-pixel SWIR sensor costs approximately \$2,000 and the DMD with its evaluation board costs approximately \$1,500, resulting in a total cost of under \$4,000 for our FPA-CS prototype. 

\section{Reconstruction algorithms}
\label{sec:CS}
\paragraph{Forward imaging model of FPA-CS.}
Our FPA-CS prototype is equivalent to an array of $64 \times 64$ SPCs working in parallel.
The DMD and sensor are synchronized to project modulation patterns and record coded low-resolution images at a certain frame rate, say $f_{DMD}$ fps; we used $f_{DMD}=480$ fps in our experiments. 

Let us describe the sensor image measured at time $t$ using the following matrix-vector notation: $y_t = A_t x_t$, where $y_t$ is a vector with sensor measurements at $4096$ pixels, $x_t$ represents the unknown high-resolution image formed at the DMD plane, and the matrix $A_t$ encodes modulation of $x_t$ with the DMD pattern and subsequent mapping onto the SWIR sensor pixels. 
Thus, $A_t$ can be decomposed into two components as $A_t = CD_t$; $D_t$ denotes a diagonal matrix that contains the binary pattern projected on the DMD at time $t$; and every $C_{i,j}$ entry in $C$ represents the response of a unit light from $j$th DMD mirror onto $i$th sensor pixel.
$C$ is a highly sparse matrix, because only a small group of roughly $16 \times 16$ DMD mirrors map to a single pixel on the SWIR sensor. Therefore, for a stationary camera assembly, we can estimate $C$ using a separate, one-time calibration procedure, which is used in all subsequent experiments. 

To reconstruct video at a desired frame rate, say $F_{r}$ fps, we divide low-resolution sensor images, $y_t$, into sets of $T = f_{DMD}/F_r$ measurements, and assume that all of them correspond to the same high-resolution image. 
Suppose the $k$th set correspond to $y_t = A_tx_t$ for $t=(k-1)T+1,\ldots,kT$; we assume that $x_t = \bfx_k$ and stack all the $y_t$ and $A_t$ in the $k$th set to form a larger system of equations. For instance, the system for $k=1$ can be written as 
\begin{equation}
\label{eq:csmodel_T_frames}
\begin{bmatrix}
y_1           \\
y_2  \\
\vdots   \\
y_T         
\end{bmatrix} = 
\begin{bmatrix}
A_1x_1           \\
A_2x_2  \\
\vdots   \\
A_Tx_T         
\end{bmatrix}
\equiv \begin{bmatrix}
	A_1           \\
	A_2  \\
	\vdots   \\
	A_T         
\end{bmatrix}\bfx_1 \; \Rightarrow\; \boxed{\bfy_k = \bfA_k \bfx_k.}
\end{equation}
Our goal is to reconstruct the $\bfx_k$ from the noisy and possibly under-determined sets of linear equations $\bfy_k = \bfA_k \bfx_k$.  

\paragraph{Total variation-based reconstruction.}

Natural images have been shown to have sparse gradients.
For such signals, one can solve an optimization problem of the following form \cite{chambolle2004algorithm,osher2005iterative}: 
\begin{equation}\label{eq:TV}
\widehat{\bf x} = \arg\min_{\bf x}\,\textrm{TV}(\bf x)\quad \text{subject to}\,\,\| \bf y - A \bf x \|_2 \le \epsilon,
\end{equation}
where the term $\textrm{TV}(\bf x)$ refers to the total-variation  of $\bfx$. 
In the context of images where~$\bfx$ denotes a 2D signal, the operator $\textrm{TV}(\bfx)$  can be defined as
\begin{align*}
\textrm{TV}(\bfx) = \sum_i \sqrt{ (D_u \bfx(i))^2 + (D_v \bfx(i))^2 },
\end{align*}
where $D_u \bf x$ and $D_v\bf x$ are the spatial gradients along horizontal and vertical dimensions of $\bf x$, respectively. 
In the context of video signals, we can also exploit the similarity between adjacent frames along the temporal direction. We can view a video signal as a 3D object that consists of a sequence of 2D images, and we expect the signal to exhibit sparse gradients along spatial and temporal directions. Thus, we can easily extend the definition of TV operator to include gradients in the temporal d
direction.
\begin{align*}
\textrm{TV}_\text{3D}(\bfx) = \sum_i \sqrt{ (D_u \bfx(i))^2 + (D_v \bfx(i))^2 + (D_t \bfx(i))^2},
\end{align*}
where $D_t \bf x$ represents gradient along the temporal dimension of $\bf x$. 
In our experimental results, we used TVAL3 \cite{li2009tval3} for the reconstruction of images and MFISTA \cite{beck2009fast} for the reconstruction of videos.


\section{Experiments}
\label{sec:expts}
To demonstrate the performance of our device, we show results on several static and dynamic scenes captured using our prototype SWIR FPA-CS camera.

\begin{figure*}[t]
	\centering
	\includegraphics[width=1\textwidth, keepaspectratio]{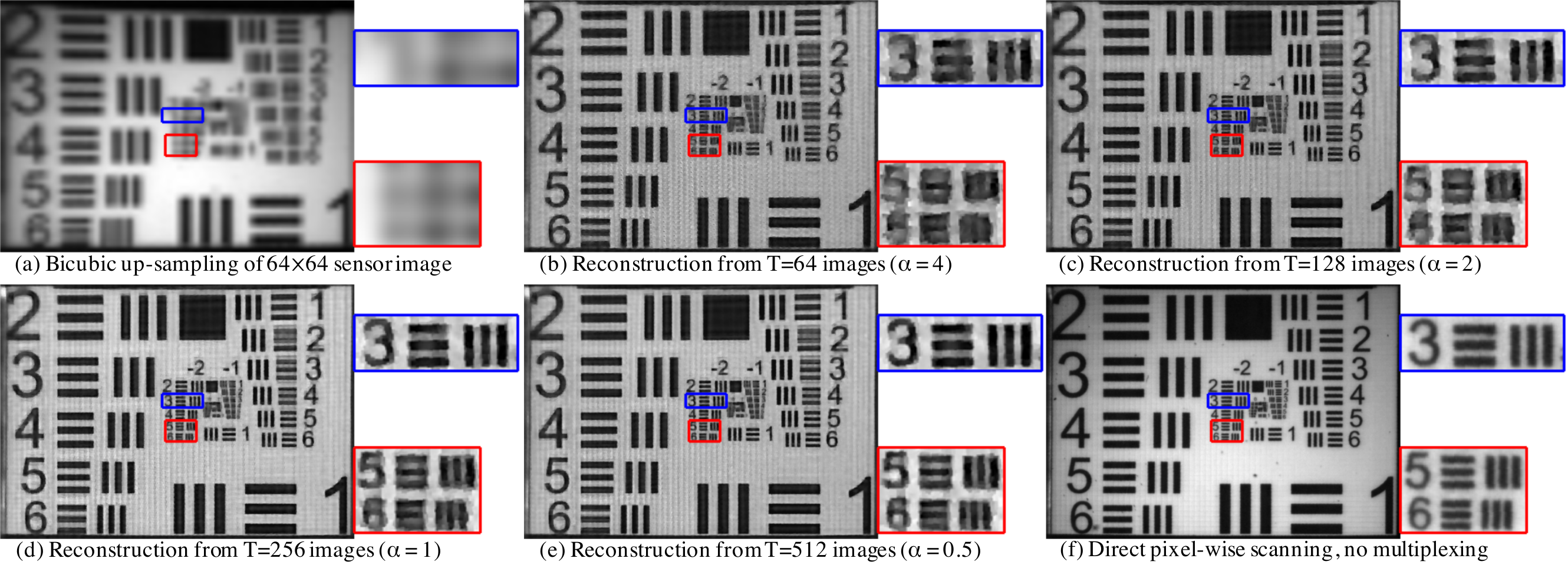}
	\caption{FPA-CS results for a resolution chart. (a) Interpolation of a $64\times64$ sensor image. (b)--(e) Images reconstructed using increasing number of measurements ($T$); compression factor $\alpha \approx 10^6/(T\times4096)$; a larger value of $T$ would translate to smaller frame rate of reconstruction for a dynamic scene. (f) Direct pixel-wise scanning result, where no multiplexing is performed and $T\approx 3000$.}
	\label{fig:resolution_charts}
\end{figure*}

\vspace{-4mm}
\paragraph{Resolution chart.} 
To study the spatial resolution characteristics of our system, we first captured images of a USAF 1951 target using our prototype device.
For each measurement, the DMD projected a random binary spatial pattern and the sensor recorded a $64\times 64$ image. 
We adjust the exposure duration for each acquired images to $0.8$ ms and acquired $512$ sensor images with varying patterns on the DMD. 
We then reconstructed the scene at the same resolution as that of the DMD using the TV-regularized reconstruction algorithm described in Section~\ref{sec:CS}. 
To study the impact of the number of measurements, we reconstructed the image with $T=64, 128, 256$, and $512$ cpatured images, which 
correspond to compression of $\alpha = 4,2,1, \text{ and } 0.5$, respectively. We defined $\alpha \approx 10^6/(T\times4096)$.
Figure \ref{fig:resolution_charts} presents the results obtained at various compression rates.

Figure~\ref{fig:resolution_charts}(a) presents an up-sampled version of a single $64\times 64$ image using bicubic interpolation. 
Figure \ref{fig:resolution_charts}(b)--(e) present images reconstructed with different number of measurements; the spatial quality of reconstructed images improves as the number of measurements increases. 
Figure~\ref{fig:resolution_charts}(f) presents an image reconstructed using direct pixel-wise scanning of the DMD without multiplexing. This allows us to acquire images at the maximum spatial resolution that FPA-CS can provide (limited only by the performance of optics), at the sacrifice of temporal resolution.
To perform the direct pixel-wise scanning, we divided the DMD into $18\times 20$ regions that map to non-overlapping sensor pixels. We sequentially turned on one micromirror in all the groups and recorded the respective sensor images. In this manner, we can compute the image intensity at every micromirror location. Such pixel-wise scanning, with non-overlapping division of the DMD, requires approximately 3000 images to be captured. 
Such scanning can also be considered a super resolution scheme performed in the optical domain, but it can only be used for static scenes. 

Overall, results in Figure~\ref{fig:resolution_charts} demonstrates that high-resolution images can be obtained from a small number of multiplexed images. Furthermore, FPA-CS provides flexible tradeoff between spatial and temporal resolutions of the reconstructed signals. As we increase the number of images used for reconstruction, the spatial quality improves, but the imaging interval per frame also increases. Therefore, a small number of multiplexed images can be used to reconstruct dynamic scenes at high temporal resolution, or static scenes can be reconstructed using a large number of multiplexed images.

\begin{figure*}[t]
	\centering
	\includegraphics[width=1\textwidth, keepaspectratio]{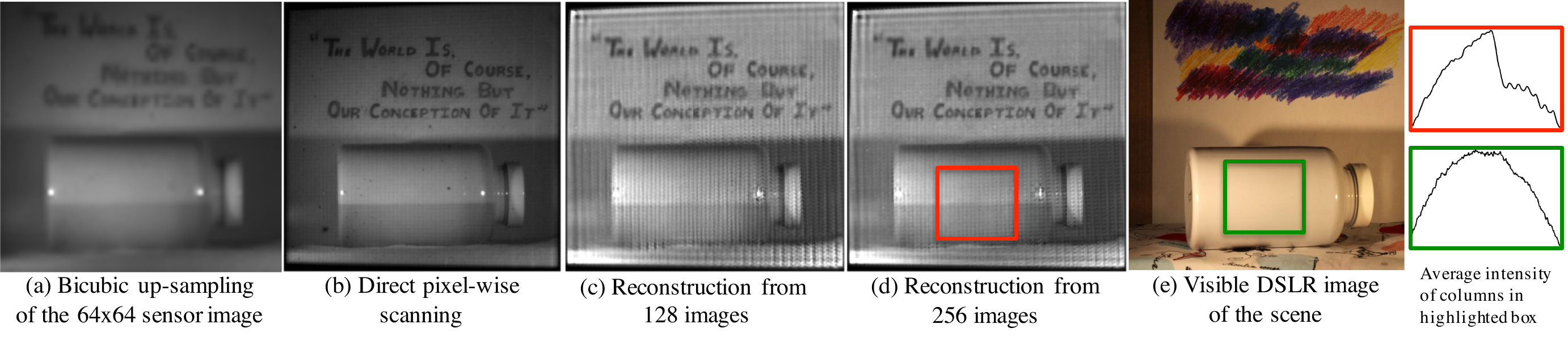}
	\caption{SWIR imaging of an opaque bottle in the foreground and a crayon scribble in the background. (a) Bicubic interpolated version of a $64\times64$ sensor image. (b) Direct pixel-wise scanning result. (c) and (d) Images reconstructed with different compression factor $\alpha$. (e) Visible image of the scene taken with a DSLR camera. Notice that the crayons are transparent in SWIR allowing us to read the text behind the  scribbles. The bottle in the foreground is opaque in visible but transparent in SWIR. Highlighted boxes on the right display mean intensity of columns in the corresponding regions of the SWIR and visible images, showing that one can estimate the water level inside the bottle from the SWIR image, but not from the visible image.}
	\vspace{-0.2in}
	\label{fig:Invisible}
\end{figure*}

\vspace{-4mm}
\paragraph{Highlighting capabilities of imaging in SWIR.} We  present  a simple experiment that highlights two attributes of SWIR imaging.
The scene in Figure \ref{fig:Invisible}, when observed in the visible spectrum, is largely unremarkable and consists of an opaque bottle and a crayon scribble in the background.
However, the corresponding SWIR images show two interesting differences.
First, note that the crayon scribble is transparent in SWIR, therefore allowing us to read the text behind it.
Second, the bottle is partially transparent in SWIR, therefore allowing us to see through the bottle and determine the water level inside the bottle.
Figure \ref{fig:Invisible} also shows 1D plots of the mean intensity of columns inside the highlighted color boxes; SWIR intensity changes because of the water inside the bottle, but the visible light intensity remains unchanged.

\begin{figure*}[t]
	\centering
	\includegraphics[width=1\textwidth, keepaspectratio]{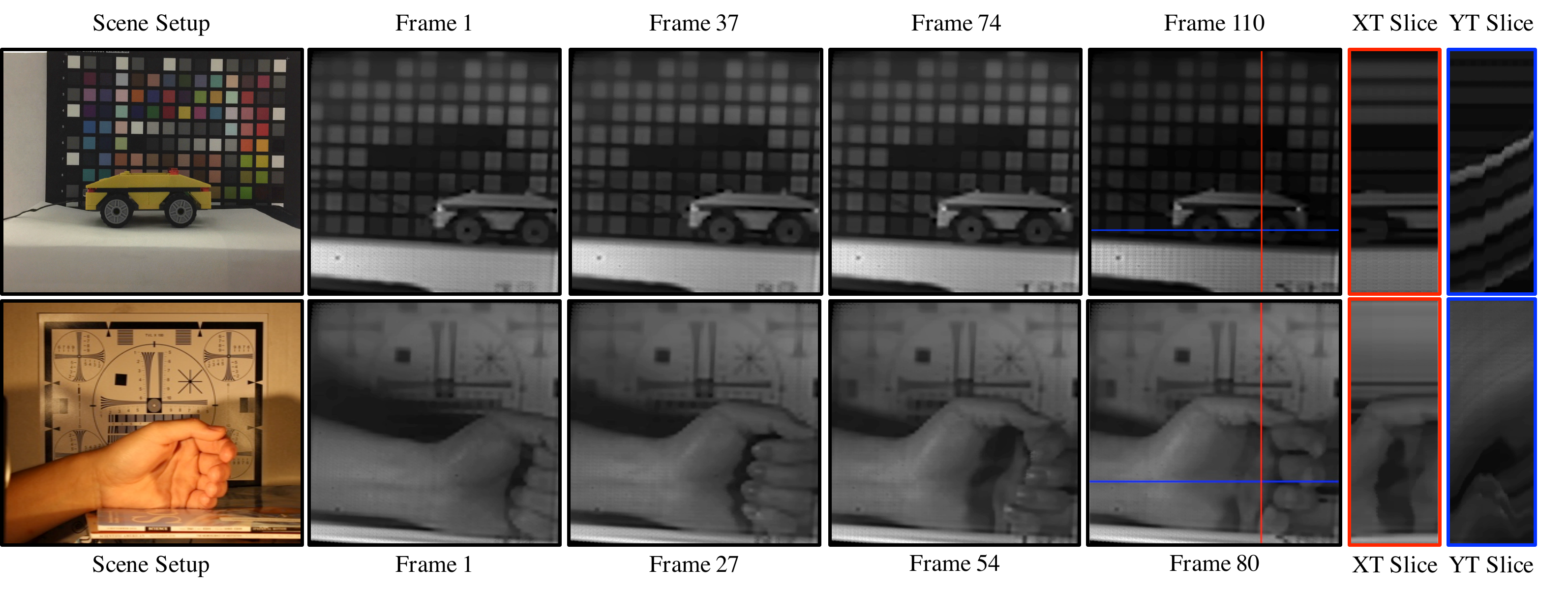}
	\caption{Selected frames from reconstructed SWIR videos. Each frame in the moving car videos is reconstructed using $T=16$ captured images ($\alpha = 16$) at 32-fps. Each frame in the moving hand videos is reconstructed using $T=22$ captured images ($\alpha = 11.6$) at 21.8-fps. Both videos are reconstructed using 3D-TV prior. XT and YT slices for both videos are shown to the right of the images.}
	\vspace{-0.2in}
	\label{fig:car_xyt}
\end{figure*}

\vspace{-4mm}
\paragraph{Video reconstruction.}
Figure \ref{fig:car_xyt} shows results on two dynamic scenes. In the moving car and moving hand video, we grouped the captured sequence  into sets of $T=16$  and $T = 22$ images, respectively. 
We used one such set to represent each frame of the video according to \eqref{eq:csmodel_T_frames};  this correspond to videos at 32  fps and 21.8 fps, with  compression factors $\alpha = 16$ and $\alpha = 11.6$, respectively. We used 3D TV optimization problem described in \eqref{eq:TV} for the reconstruction. Complete videos can be found in the supplemental material. 
\Salnote{Let's also look at videos for 16, 22, 32 fps results, which correspond to $T=32, 16, 8$ low-res images per high-res image... maybe we can add some discussion on the tradeoffs between spatial/temporal resolution}

\section{Discussions}
\paragraph{Artifacts.}
Some artifacts can be observed in the output images and videos from the FPA-CS system. The two major types of artifacts that can be observed are: (1) motion artifacts that occur at sharp edges of moving objects in the scene, and  (2) ``blocky'' structural artifacts that underlie the entire captured scene. 
The motion artifacts are simply an artifact of motion blur while
the ``blocky'' artifacts are caused by small misalignments in the mapping between the sensor pixel and DMD pixel introduced after the system calibration.
In practice, we observed that applying a 3D median filter can largely suppress both artifacts. 

\paragraph{Choice of modulation masks.}
To obtain high fidelity reconstruction, modulation masks should meet two conditions.
First, the system matrix in \eqref{eq:csmodel_T_frames}---the combined system of 4096 SPCs in every set of $T$ frames---should be well-conditioned so that the image reconstruction process is stable and robust to noise. 
Second, the spatial code should have high light-throughput that maximizes the signal-to-noise ratio in sensor images.
In this paper, we tested two mask patterns--Hadamard and random binary--both of which satisfy these characteristics.
%
Hadamard matrices are known to be optimal in terms of linear multiplexing  \cite{schechner2007multiplexing}.
The results shown in Figure~\ref{fig:Invisible} correspond to Hadamard measurements.
We also used random binary patterns since they are known to satisfy the restricted isometry property \cite{candes2008restricted}, and therefore lead to robust inversion when used along with sparse regularizers such as total variation.
The results shown in Figure~\ref{fig:resolution_charts} and \ref{fig:car_xyt} correspond to binary random measurements.
Notice that in both cases, $50$\% of the light reaches the sensor after modulation.
In practice, we observed that the reconstructions obtained  from the two modulation patterns were near-identical.

\vspace{-4mm}
\paragraph{Benefits.}
FPA-CS provides three advantages over conventional imaging.
First, our CS-inspired FPA-CS system provides an inexpensive alternative to achieve high-resolution SWIR imaging.
Second, compared to traditional single pixel-based CS cameras, FPA-CS simultaneously records data from $4096$ parallel,  compressive systems, thereby significantly improving the measurement rate.
As a consequence, the achieved spatio-temporal resolution of our device is orders of magnitude better than the SPC.
 
\vspace{-4mm}
\paragraph{Limitations.}
FPA-CS exploits spatio-temporal redundancy in the reconstruction, therefore, extremely complex scenes such as a bursting balloon cannot be directly handled by the camera.
Since the spatio-temporal redundancy exploited by traditional compression algorithms and our imaging architecture are very similar, one can assume that the scenes that can be compressed efficiently, can also be processed well using our method. 
Our prototype uses a binary per-pixel shutter, which causes a $50\%$ reduction in light throughput as half of the light is wasted.
In future, a separate, synchronized $64 \times 64$ image sensor can be used in the other arm, thereby doubling the measurement rate and further increasing the spatio-temporal resolution that can be achieved.
The algorithm is currently not real-time and thus precludes the direct-view capability.

\vspace{-4mm}
\paragraph{Conclusion.}
We presented focal plane array-compressive sensing (FPA-CS), a new imaging architecture for parallel compressive measurement acquisition that can provide quality videos at high spatial and temporal resolutions in SWIR. 
The architecture proposed here is generic and can be adapted to other spectral regimes such as mid-wave infra-red and thermal imaging, where, much like SWIR, sensors are prohibitively expensive.

\vspace{-4mm}
\paragraph{Acknowledgements:} This work was partially supported by NSF Grants IIS:1116718 and CCF:1117939. Early versions of this work was supported by the Office of Naval Research.
The authors wish to thank Richard G. Baraniuk, Jason Holloway, Kevin Kelly, Bob Muise and Adam Samaniego for their feedback and discussions on the early versions of this project.

\vspace{-2mm}

{\small
\bibliographystyle{ieee}
 \nocite{wagadarikar2008single}
\bibliography{FPA-CS,videocs}
}

\end{document}